\documentclass{article}
\usepackage{stdlib}

\title{Interactive Robot Transition Repair With SMT}

\author{
Jarrett Holtz,
Arjun Guha, and
Joydeep Biswas
\\
University of Massachusetts Amherst \\
\{jaholtz,arjun,joydeepb\}@cs.umass.edu
}

\begin{document}

\maketitle

\begin{abstract}
Complex robot behaviors are often structured as state machines, where states
encapsulate actions and a transition function switches between states. Since
transitions depend on physical parameters, when the environment
changes, a roboticist has to painstakingly readjust the parameters to
work in the new environment.
We present \emph{interactive SMT-based Robot Transition Repair} (\technique):
instead of manually adjusting parameters, we ask the roboticist to identify a
few instances where the robot is in a wrong state and what the right state
should be. An automated analysis of the transition
function 1) identifies adjustable parameters, 2)
converts the transition function into a system of logical constraints,
and 3) formulates the constraints and user-supplied corrections as a \maxsmt{} problem that
yields new parameter values.
We show that \technique{} finds
new parameters 1) quickly, 2) with few corrections, and 3) that the parameters generalize
to new scenarios. We also show that a \technique{}-corrected state machine
can outperform a more complex, expert-tuned state machine.

\end{abstract}

\section{Introduction}

Complex robot control software is typically structured as a state machine,
where each state encapsulates a feedback controller.
Even if each state is correct, the transitions between states depend
on parameters that are hard to get right, even for experienced
roboticists.
It is very common for parameter values to work in simulation
but fail in the real world, to work in one physical environment but fail
in another, or to work on one robot but fail on another.
For example, \figref{attacker-examples} shows the trajectory of a
robotic soccer player as it tries to kick a moving ball. A very small change
to its parameter values determines whether or not it succeeds.

Even a simple robot may have a large parameter space, which makes
exhaustive-search impractical. Moreover, robot performance is usually
non-convex with respect to parameter values, which makes general optimization
techniques susceptible to local minima.
For some cases, there exist calibration procedures to adjust
parameters automatically (\eg{} \cite{holtz2017automatic}),
but these are not general procedures.
Therefore, roboticists usually resort to adjusting parameters
manually---a tedious task that can result in poor performance.

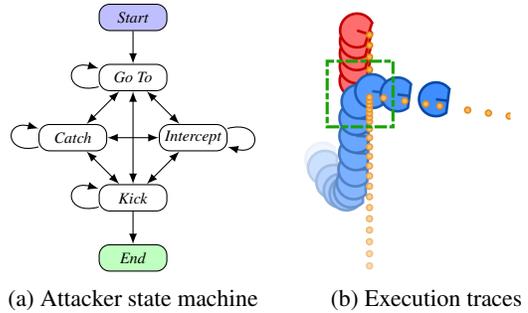
\begin{figure}[t]
 \centering
 \begin{subfigure}[b]{0.49\linewidth}
 \centering
 \begin{tikzpicture}[scale=0.05]
  \tikzstyle{every node}+=[rounded corners,minimum width=0.9cm,minimum
  height=0.35cm,node distance=0.8cm and 1.7cm,draw=black,inner
sep=0pt,font=\tiny\itshape]
  \tikzstyle{every path}+=[>=latex]
  \node[fill=blue!25] (start) {Start};
  \node (goto) [below of=start] {Go To};
  \coordinate[below of=goto] (belowgoto);
  \node (intercept) [right of=belowgoto] {Intercept};
  \node (catch) [left of=belowgoto] {Catch};
  \node (kick) [below of=belowgoto] {Kick};
  \node[fill=green!25] (end) [below of=kick] {End};
  \path[every loop/.style={looseness=5}]
    (start) [->] edge (goto)
    (goto) [<->] edge (kick)
    (goto) [<->] edge (intercept)
    (goto) [<->] edge (catch)
    (intercept) [<->] edge (kick)
    (catch) [<->] edge (kick)
    (catch) [<->] edge (intercept)
    (kick) [->] edge (end)
    (goto) [->] edge [loop left] (goto)
    (catch) [->] edge [loop left] (catch)
    (intercept) [->] edge [loop right] (intercept)
    (kick) [->] edge [loop left] (kick);
\end{tikzpicture}
 \caption{Attacker state machine}
  \figlabel{example-model}
 \end{subfigure}
 \begin{subfigure}[b]{0.4\linewidth}
  \includegraphics[width=0.85\linewidth]{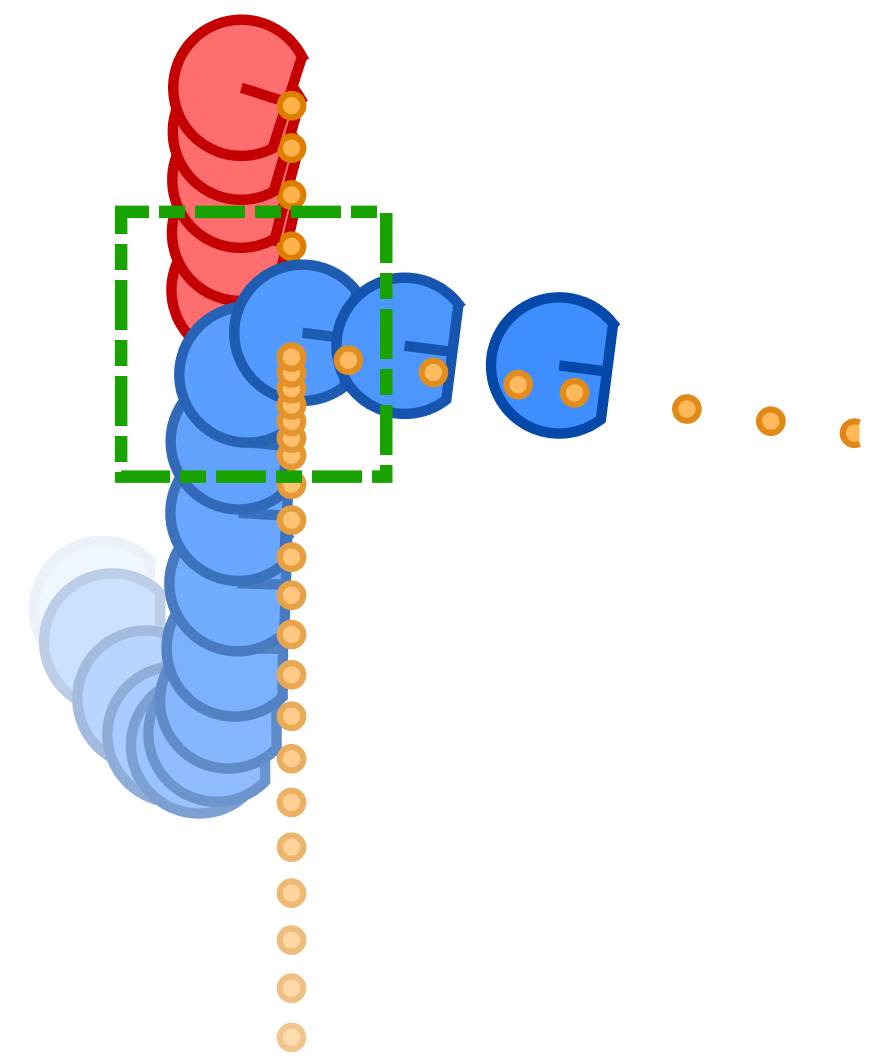}
  \caption{Execution traces}
  \figlabel{example-trajectories}
 \end{subfigure}
 \caption{A robot soccer attacker a) state machine, with b)
   successful (blue) and unsuccessful (red) traces to {\itshape Intercept}
   a ball (orange) and {\itshape Kick} at the goal. The green box isolates the error: the
successful trace transitions to {\itshape Kick}, the unsuccessful trace remains
in {\itshape Intercept}.}
  \figlabel{attacker-examples}
\end{figure}

However, consider the following observations: a roboticist debugging a robot
can frequently identify when something goes wrong,
and what the robot should have done instead. When the robot control software
is structured as a state machine, this corresponds to identifying when
the robot is in the wrong state and what the correct state should be.
This is a \emph{partial specification} of expected behavior: \ie{} it does not
enumerate a complete sequence of states, the parameters to
adjust, how to adjust them, or even identify all errors. In this
paper, we show that partial specification of this kind are adequate inputs
for an automatic parameter repair procedure.

We present \emph{Interactive SMT-based Robot Transition Repair} (\technique{}).
During execution,
\technique{} logs the execution trace of the transition function. After execution,
the roboticist examines the trace and provides a handful of corrections.
\technique{} then converts the set of corrections and the
transition function into a logical formula for a \maxsmt{} solver.
A solution to this \maxsmt{} problem minimizes adjustments to the parameters,
while maximizing the number of satisfied corrections.

Our experimental results demonstrate that \technique{}
\begin{inparaenum}[1)]
\item is more computationally tractable than exhaustive parameter search;
\item is scalable;
\item generalizes to unseen situations; and
\item when applied to simple robot soccer attacker, can
  outperform a more complicated, expert-tuned attacker that won the
  lower bracket finals at \emph{RoboCup 2017}.
\end{inparaenum}

\section{Related Work}
\seclabel{related}

Finite state machines (FSMs) can specify robot behavior and enable robust
autonomy. RoboChart~\cite{miyazawa2017} uses FSMs to verify liveness,
timeliness, and other properties of robots. The HAMQ-INT
algorithm~\cite{bai2017efficient} uses
reinforcement learning to discover internal
transitions and adapt to unexpected scenarios. \technique{} also leverages FSMs,
but focuses on correcting transition errors that occur when the robot is deployed
in a new environment.

Robot behavior can also be repaired by dynamic synthesis of new
control structures, such as automatic synthesis of new
FSMs~\cite{wong2014correct}, or synthesis of code from a context-free motion
grammar with parameters derived from human-inspired control~\cite{dantam2013}.
When
automated synthesis is intractable, a user-generated specification
in a domain-specific language can be used to synthesize
a plan~\cite{nedunuri2014}.
\technique{} assumes that the FSM does not need new states or transitions,
but that failures are due to \emph{incorrect triggering} of the transition
function arising from incorrect environment-dependent parameter values.

Robot behaviors rely on environment-dependent parameters for robust
and accurate execution. If a precise model of the dependency between parameters and
behaviors is available, it may be possible to design
a \emph{calibration procedure} that executes a specific sequence of actions and
to recover correct parameter
values~(\eg{}\cite{holtz2017automatic}). If a calibration procedure
cannot be designed, but the effect of parameters
is well-understood, it may be possible to optimize for the parameters using a
functional model~\cite{cano2016automatic}.
Model-based diagnosis can diagnose faulty parameters~\cite{reiter1987theory} if
the behavior of the robot in its environment can be formally defined.
\technique{} can repair parameters even without a descriptive model of the
robot's behavior.

Human input can help overcome the limitations of autonomous
algorithms~\cite{Kamar2016DirectionsIH,nashed2018human}. Learning from
demonstration (LfD)~\cite{argall2009survey} and inverse reinforcement
learning (IRL)~\cite{abbeel2011inverse} allow robots to learn new behaviors
from human demonstrations. LfD can also overcome
model errors by correcting portions of the state
space~\cite{mericli2012multi}. These approaches require demonstrations
in the full high-dimensional state space of the robot,
which can be tedious for users to provide.
When human demonstration does not specify
\emph{why} an action was applied to a state, it can be hard to generalize
to a new situation. \technique{} generalizes to
new scenarios since it infers dependencies from the code
of the transition function. Furthermore, it requires only a partial
specification of correct behavior.

Domain-specific languages (DSLs) allow users to
specify high-level behavior using abstractions such as Instruction
Graphs~\cite{mericli2014interactive}. Unlike DSLs that provide a complete
specification of robot behavior, \technique{} only requires sparse corrections
that partially-specify expected behavior.

DirectFix~\cite{mechtaev:directfix} formulates program repair as a \maxsmt{} problem
and deems a program fixed when all tests pass.
Physical robots don't have deterministic test cases
and user-provided corrections can be contradictory.
\technique{} uses \maxsmt{} to minimize changes and maximize
the number of satisfied corrections.
There are other domains that are not amenable to unit-testing.
For example, Tortoise~\cite{tortoise:weiss} propagates system
configuration fixes from the shell to a system configuration
specification. However, Tortoise requires a complete fix where
\technique{} only requires a partial specification.

\emph{Programming by example} synthesizes programs from
a small number of examples~\cite{gulwani2011automating} and can also
support noisy data~\cite{devlin:robustfull}
\emph{Program templates} can help  synthesize task and motion
plans~\cite{nedunuri2014}.
\technique{} does not synthesize new program structure, but focuses
on making minimal parameter adjustments using \maxsmt{} to satisfy user
corrections.

\section{Background}
\seclabel{background}

We use a real-world example to motivate \technique: a robot soccer attacker
that
\begin{inparaenum}[1)]
\item goes to the ball if the ball is stopped,
\item intercepts the ball if it is moving away from the attacker,
\item catches the ball if it is moving toward the attacker, and
\item kicks the ball at the goal once the attacker has control of the ball.
\end{inparaenum}
Each of these sub-behaviors is a distinct, self-contained feedback
controller (\eg{} ball interception~\cite{biswas2014opponent} or
omnidirectional time optimal control~\cite{kalmar2004near}).
At each time-step,
the attacker 1) switches to a new controller if
necessary and 2) invokes the current controller
to produce new outputs. We represent the attacker
as a \emph{robot state machine} (RSM), where each state represents a
controller (\figref{example-model}).

In this paper, we assume that the output of each controller is
nominally correct: there may be minor \emph{performance} degradation
when environmental factors change, but we assume that they are
\emph{convergent}, and will eventually produce the correct result.
However, environmental factors also affect the transition function, which
transfers execution from one controller to another. For example, the friction
coefficient between the ball and the carpet
affects: when the attacker transitions from {\itshape Intercept}
to {\itshape Kick}; and the mass of the ball affects when the
attacker transitions from {\itshape Kick} to {\itshape Done}.
These factors vary from one environment to another.
Since transition functions do not have any
self-correcting mechanisms, robots are prone to behaving incorrectly when their
parameters are incorrect for the given environment.

\subsection{Robot State Machines}

A robot state machine (RSM) is a discrete-time Mealy machine that is extended
with continuous inputs, outputs, and program variables. Formally, an RSM is a
9-tuple $\langle S, S_0, S_F, V, V_0, Y, U, T, G\rangle$, where $S$ is the
finite set of states, $S_0 \in S$ is the start state, $S_F \in S$ is the end
state, $V \in \reals{m}$ is the set of program variable values, $V_0 \in
\reals{m}$ are the initial values of the program variables, $Y\in\reals{n}$ are
the continuous inputs, $U\in\reals{l}$ are the continuous actuation outputs, $T
: S \times Y \times V \rightarrow S$ is the transition function, and $G : S
\times Y \times V \rightarrow U \times V$ is the emission function. At each time
step $t$, the RSM first uses the transition function to select a state and
then the emission function to run the controller associated with that state.
The transition function can only update the current state, whereas the emission
function can update program variables and produce outputs.


\subsection{Transition Errors}

\figref{example-trajectories} shows two traces of the attacker. In the
blue trace, the attacker correctly intercepts the moving ball and kicks it at
the
goal. But, in the red trace, the attacker fails to kick: it remains stuck
in the \emph{Intercept} state and never transitions to \emph{Kick}.
Over the course of several trials (\eg{} a robot soccer game), we may find that
the attacker only occasionally fails to kick.
When this occurs, it is usually the case that the high-level structure of
the transition function is correct, but that the values of the
parameters need to be adjusted. Unfortunately, since there are 11 real-valued
parameters in the full attacker RSM, the search space is large.

To efficiently search for new parameter values, we need to reason
about the structure of the transition function. To do so,
the next section describes how we systematically convert
it to a formula in propositional logic, extended
with arithmetic operators. This formula encodes the structure of the
transition function along with constraints from the user corrections, and allows
an SMT solver to efficiently find new parameters to correct the errant
transition(s).

\section{Interactive Robot Transition Repair}
\seclabel{repairable_controllers}

The \technique{} algorithm has four inputs: 1) the transition function,
2) a map from parameters to their values, 3) an
execution trace, and 4) a set of user-provided corrections.
The result of \technique{} is a corrected parameter map that maximizes
the number of corrections satisfied and minimizes the changes to the
input parameter map. (The tradeoff between these objectives is a
hyperparameter.)

\technique{} has three major steps.
\begin{inparaenum}[1)]
  \item For each user-provided correction,
  it \emph{partially evaluates}
  the transition function for the inputs and variable
  values at the time of correction, yielding \emph{residual transition functions}
  (\secref{program-analysis}).
  \item Finally, it uses the residual transition functions to formulate an
  optimization problem
  for an off-the-shelf \maxsmt{} solver, for this paper we use
  Z3~\cite{bjorner2015nuz}. The solution to
  this  problem is an adjustment to the parameter values (\secref{max-smt}).
\end{inparaenum}

To illustrate the \technique{} algorithm, we present as a running example a simplified
attacker RSM that is only capable of handling
a stationary ball on the field (\figref{simple-rsm}).
Therefore, the RSM has four
states (Start, GoTo, Kick, and End) and its transition function
(\figref{example-function}) has
four parameters (\cpparam{aimMargin}, \cpparam{maxDist}, \cpparam{viewAng}, and
\cpparam{kickTimeout}). From an execution trace of the RSM (\figref{example-inputs}), we
consider an example
where at time-step $t=5$, the user identifies that the transition function produces an
incorrect result: instead of GoTo, it should have returned Kick. With this
example in mind, we present how \technique{} uses the transition function code, an
 execution trace, and a correction to
identify that an adjustment to just one of the parameters, \cpparam{maxDist}, is sufficient
to satisfy this correction (\figref{example-outputs}).

\subsection{Transition Functions and \technique{} Inputs}
\seclabel{syntax}

To abstract away language-specific details of our repair procedure, we present
\technique{} for an idealized imperative language that only has features
essential for writing transition functions.  \figref{syntax} lists the
syntax for a  transition function written in \technique{}-repairable form using
a notation that is close to standard BNF.  In
general, the transition function consists of sequences of statements comprised
of expressions over 1) the current state \pstate{}, 2) program inputs \pin{x},
3) parameters \pparam{x}, and 4) program variables \pvar{x}. Based on
computations over these identifiers, the transition function returns the next
state.  \figref{syntax} has a list of operators that often appear in transition
functions, such as arithmetic and trigonometric functions, but the list is
\emph{not} exhaustive. As a concrete example of a transition function written in
repairable form, \figref{example-function} shows the transition function
for the running example: it branches on the current state (\pstate) and returns
the next state. The crux of the transition function are the conditions that
determine when the transition from \state{Go To} to \state{Kick} occur.

\begin{figure}[t]
  \centering
  \begin{tikzpicture}[scale=0.05]
  \tikzstyle{every node}+=[rounded corners,minimum width=0.9cm,minimum
  height=0.35cm,node distance=1.5cm and 1.7cm,draw=black,inner
  sep=0pt,font=\tiny\itshape]
  \tikzstyle{every path}+=[>=latex]
  \node[fill=blue!25] (start) {Start};
  \node (goto) [right of=start] {Go To};
  \node (kick) [right of=goto] {Kick};
  \node[fill=green!25] (end) [right of=kick] {End};
  \path[every loop/.style={looseness=15}]
  (start) [->] edge (goto)
  (goto) [->] edge [loop below] (goto)
  (goto) [->] edge (kick)
  (kick) [->] edge [loop below] (kick)
  (kick) [->] edge (end);
  \end{tikzpicture}
  \caption{Simplified attacker RSM.}
  \figlabel{simple-rsm}
\end{figure}
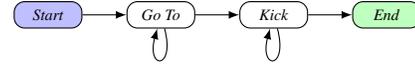

A parameter map ($P$) specifies the parameters of a transition function.
\figref{example-inputs} shows the parameter map of our example before running
\technique{}. The output of \technique{} will be a set of adjustments to this
parameter map.

An execution trace is a sequence of \emph{trace elements} $\trelt_t$. A
trace element records the values of sensor inputs, program variables, and the
state at the start of time-step $t$. Finally, a user-provided \emph{correction}
($\correlt{}$) specifies the expected state at the end of time-step $t$.
In our example, the attacker should have transitioned to the \state{Kick}
state after $t=5$.

\figref{example-inputs} shows the trace element and a user correction for the
running example: since the correction $c_5$ refers only to the time-step $t=5$,
only the relevant trace element $\trelt_5$ is shown.

\begin{figure*}%
\begin{subfigure}[b]{0.355\textwidth}
\lstset{language=srtr-example}
\scriptsize
\scriptsize
\textbf{Transition function} ($T$)
\begin{lstlisting}
if ($\pstate$ == "START") {
  return "GOTO";
} else if ($\pstate$ == "GOTO") { #\label{line:goto-start}#
  $\cpid{relLoc}$ := $\cpin{ballLoc} - \cpin{robotLoc}$;  #\label{line:relLoc}#
  $\cpid{aimErr}$ := $\textrm{AngleMod}(\cpin{targetAng} - \cpin{robotAng})$;
  $\cpid{robotYAxis}$ := $\langle \sin(\cpin{robotAng}), \psub\!\cos(\cpin{robotAng})\rangle$; #\label{line:trig-input}#
  $\cpid{relLocY}$ := $\cpid{robotYAxis} \cdot \cpid{relLoc}$;
  $\cpid{maxYLoc}$ := $\cpparam{maxDist} \cdot \sin(\cpparam{viewAng})$; #\label{line:maxYLoc}#  #\label{line:trig-param}#
  if ($\cpid{aimErr} < \cpid{aimMargin} \wedge \left\lVert\cpid{relLoc}\right\rVert < \cpparam{maxDist} \wedge$  #\label{line:kick-guard-start}#
      $\left\lVert\cpid{relLocY}\right\rVert < \cpid{maxYLoc} \wedge$
      $\cpin{time} > \cpvar{lastKick} + \cpparam{kickTimeout}$) {  #\label{line:kick-guard-end}#
    return "KICK";
  } else return "GOTO"; #\label{line:goto-end}#
} else if ($\pstate == "KICK" \wedge $
           $\cpvar{timeInKick} > \cpparam{kickTimeout}$) {
  return "END";
} else return "KICK";
\end{lstlisting}
\caption{A simple RSM and its transition function.}
\figlabel{example-function}
\end{subfigure}
\vrule\;
\begin{subfigure}[b]{0.23\textwidth}
\lstset{language=srtr-example}
\scriptsize
\(
\begin{array}{@{}r@{\;}c@{\;}l}
\multicolumn{3}{@{}l}{\textbf{Parameter map ($P$)}} \\
P & = & \langle \cpparam{aimMargin} \pmapsto \pi/50, \\
& & \phantom{\langle} \cpparam{maxDist} \pmapsto 80, \\
& & \phantom{\langle} \cpparam{viewAng} \pmapsto \pi/6, \\
& & \phantom{\langle} \cpparam{kickTimeout} \pmapsto 2\rangle\\[0.7em]
\hline
\multicolumn{3}{@{}l}{\textbf{Trace element ($\trelt_5$)}} \\
\treltins{\trelt_5} & = &
  \langle \cpin{ballLoc} \pmapsto \langle 30,40 \rangle, \\
  & & \phantom{\langle} \cpin{robotLoc} \pmapsto \langle 0,0 \rangle, \\
  & & \phantom{\langle} \cpin{robotAng} \pmapsto 0, \\
  & & \phantom{\langle} \cpin{targetAng} \pmapsto \pi/60, \\
  & & \phantom{\langle} \cpin{time}\pmapsto 5\rangle\\
\treltvars{\trelt_5} & = & \langle \cpvar{lastKick} \pmapsto 2 , \\
  & & \phantom{\langle} \cpvar{timeInKick}\pmapsto 0\rangle \\
\treltstate{\trelt_5} & = & \texttt{"GOTO"} \\[0.7em]
\hline
\multicolumn{3}{@{}l}{\textbf{Trace ($\mathcal{R}$)}} \\
\mathcal{R} & = & \langle \cdots \tau_5 \cdots \rangle \\[0.7em]
\hline
\multicolumn{3}{@{}l}{\textbf{Correction ($c_5$)}} \\
c_5 & ::= &  s_6 \pmapsto \texttt{"KICK"}
\end{array}
\)
\caption{Inputs to \technique{}.}
\figlabel{example-inputs}
\end{subfigure}
\;
\vrule
\;
\begin{subfigure}[b]{0.38\textwidth}
\lstset{language=srtr-example}
\scriptsize

\textbf{Repairable and unrepairable parameters}

\(
\begin{array}{@{}r@{\,}c@{\,}l}
\textsf{Rep}(T) & = & \{ \cpparam{aimMargin}, \cpparam{maxDist},
  \cpparam{kickTimeout} \} \\
\textsf{Unrep}(T) & = & \{ \cpparam{viewAng} \}
\end{array}
\)

\vskip 0.5em
\hrule
\vskip 0.2em
\textbf{Result of } $\textsf{MakeResidual}(T,\tau_5,P)$
\vskip -0.7em
\begin{lstlisting}
if ($\pi/60 < \cpparam{aimMargin} \wedge  50 < \cpparam{maxDist} \wedge$
    $40 < \cpparam{maxDist} \cdot 0.5 \wedge 5 > 2 + \cpparam{kickTimeout}$) {
  return "KICK";
} else return "GOTO";
\end{lstlisting}

\vskip -0.5em
\hrule
\vskip 0.2em
\textbf{Result of} $\textrm{CorrectOne}(T,\tau_5,P,c_5)$:

\(
\begin{array}{@{}r@{\,}c@{\,}l}
\phi  & = & \exists \delta^1,\delta^2,\delta^3 : \pi/60 < \pi/50+\delta^1 \wedge 50 < 80+\delta^2 \wedge \\
& & \phantom{\exists\delta^1,\delta^2,\delta^3 : } 40 < (80 + \delta^2) \cdot 0.5 \wedge 5 > 2 + (2 + \delta^3)\\
\end{array}
\)

\vskip 0.5em
\hrule
\vskip 0.2em
\textbf{Result of} $\textrm{CorrectAll}(T,P,\mathcal{R},\{c_5\})$:

\(
\begin{array}{@{}r@{\,}c@{\,}l}
\Phi &  = & \exists \delta^1,\delta^2,\delta^3,w^1 : w^1 = H \veebar (w^1 = 0 \wedge \phi)
\end{array}
\)

\vskip 0.5em
\hrule
\vskip 0.2em
\textbf{Result of} $\textrm{SRTR}(T,P,\mathcal{R},\{c_5\})$ for $H = 1$:

\(
\begin{array}{@{}r@{\,}c@{\,}l}
& & \underset{w^1,\delta^{1\cdots 3}}{\mathrm{arg\,min}} w^1 + \| \delta^1\| + \| \delta^2\| + \| \delta^3\| ~\textrm{constrained by}~\Phi \\
& = & \langle w^1 = 0, \delta^1 \mapsto 0, \delta^2 \mapsto 0.5, \delta^3 \mapsto 0 \rangle \\
\end{array}
\)
\caption{Each step of the \technique{} algorithm.}
\figlabel{example-outputs}
\end{subfigure}
\caption{\technique{} applied to a simplified robot soccer attacker with a single correction.}
\end{figure*}

\begin{figure}
\scriptsize
\lstset{language=model}
\begin{minipage}[t]{0.45\columnwidth}
\(
\begin{array}{@{}r@{\;}c@{\;}ll}
\multicolumn{4}{@{}l}{\textbf{Unary Operators}} \\
\mathit{op}_1 & ::= & \multicolumn{2}{l}{\texttt{-}
  \mid \texttt{sin}
  \mid \texttt{cos}
  \mid \cdots} \\[.4em]
\multicolumn{4}{@{}l}{\textbf{Expressions}} \\
\expr & ::=  & k & \hspace{20pt} \textrm{Constants} \\
  & \mid & \pstate  & \hspace{20pt} \textrm{State} \\
  & \mid & \pvar{x} & \hspace{20pt} \textrm{Variables} \\
  & \mid & \pin{x} & \hspace{20pt} \textrm{Inputs} \\
  & \mid & \pparam{x} & \hspace{20pt} \textrm{Parameters} \\
  & \mid & \multicolumn{2}{l}{\hspace{-6pt} \mathit{op}_1(\expr)} \\
  & \mid & \multicolumn{2}{l}{\hspace{-6pt} \expr_1~\mathit{op}_2~\expr_2}
\\[.4em]
\multicolumn{4}{@{}l}{\textbf{Transition Functions}} \\
T & ::=  & \multicolumn{2}{l}{\texttt{\{ $\stmt_1$; $\cdots$; $\stmt_n$ \}}}
\end{array}
\)
\end{minipage}
\vrule
\hspace{1pt}
\begin{minipage}[t]{0.48\columnwidth}
\(
\begin{array}{@{}r@{\;}c@{\;}ll}
\multicolumn{4}{@{}l}{\textbf{Binary Operators}} \\
\mathit{op}_2 & ::= & \texttt{+}
  \mid \texttt{-}
  \mid \texttt{*}
  \mid \texttt{>}
  \mid \cdots \\[.4em]
\multicolumn{4}{@{}l}{\textbf{Statements}} \\
\stmt & ::=  & \textbf{return}~\pstate\texttt{;} \\
  & \mid & \pvar{x}~\texttt{\passign}~\expr\texttt{;} \\
  & \mid & \textbf{\texttt{if}}~(\expr)~\stmt_1~\textbf{\texttt{else}}~\stmt_2
\\
  & \mid & \texttt{\{ $m_1\cdots m_n$ \}} \\[.4em]
\multicolumn{4}{@{}l}{\textbf{Parameter Maps}} \\
P & ::= & \multicolumn{2}{l}{\langle \pparam{x}^1\pmapsto k^1\cdots
\pparam{x}^n\pmapsto k^n\rangle} \\[.4em]
\multicolumn{4}{@{}l}{\textbf{Traces}} \\
\ptrace & ::= & \lbrack \trelt_1 \cdots \trelt_n \rbrack \\[.4em]
\multicolumn{4}{@{}l}{\textbf{Corrections}} \\
\correlt & ::= &  \pstate \in S
\end{array}
\)
\end{minipage}
\vspace{0.4em}
\hrule
\vspace{0.4em}
\(
\begin{array}{@{}r@{\;}c@{\;}ll}
\multicolumn{4}{@{}l}{\textbf{Trace Elements}} \\
\trelt_t & ::= & \multicolumn{2}{@{}l}{\mktrelt{
  \langle \limforall{i=1}{m} \pin{x^i} \pmapsto k^i \rangle}
  {\langle \limforall{j=1}{n} \pvar{x^j}\pmapsto k'^j \rangle}
  {\pstate_t \pmapsto k''_s}} \\[.4em]
\end{array}
\)
\caption{Syntax of transition functions, traces, and corrections.}
\figlabel{syntax}
\end{figure}

\begin{figure}[t]
\lstset{language=algo}
\begin{lstlisting}
// Takes a transition function $T$, and returns a partially evaluated residual transition
// function $T'$ by eliminating identifiers $x^i$ using their values $k^i$.
def Peval($T$,$x^1\pmapsto k^1\cdots x^n\pmapsto k^n$); #\label{line:peval}#

// Returns the list of repairable parameters of the transition function $T$
def Rep($T$);

// Returns the list of unrepairable parameters of the transition function $T$
def Unrep($T$);

def Residual($T$,$\trelt_t$,$P$):#\label{line:residual}#
  $\mktrelt{
  \langle \limforall{i=1}{l} \pin{x^i} \pmapsto k^i \rangle}{
  \langle \limforall{j=1}{m} \pvar{x^j} \pmapsto k'^j \rangle}{
  \pstate_t \pmapsto k''_s}$ := $\trelt_t$
  $\{ \pparam{x^1},\cdots,\pparam{x^n}\}$ = Unrep($T$)#\label{line:residual-unrep}#
  $T'$ := Peval($T$,$\limforall{i=1}{l} \pin{x^i} \pmapsto k^i,
   \limforall{j=1}{m} \pvar{x^j}\pmapsto k'^j,
   \limforall{k=1}{n} \pparam{x^k}\pmapsto P(\pparam{x^k}),
   \pstate_t \pmapsto k''_s$)#\label{line:residual-input-spec}#
  return $T'$#\label{line:residual-end}#

def CorrectOne($T$,$\trelt_t$,$P$,$\correlt$):#\label{line:correctone-start}#
  $T'$ := Residual($T$,$\trelt_t$)#\label{line:correctone:residual}#
  $\{ \pparam{x^1},\cdots,\pparam{x^m}\}$ := Rep($T$)#\label{line:correctone:rep}#
  return $\exists \delta^1,\cdots,\delta^m:\correlt = T'(
\pstate_1,\pparam{x'^1} +
\delta^1,\cdots,\pparam{x'^m}+\delta^m)$#\label{line:correctone-end}#

def CorrectAll($T$, $P$, $\ptrace$, $\{\correlt^1, \cdots, \correlt^n\}$):
  $\{ \pparam{x^1},\cdots,\pparam{x^m}\}$ = Rep($T$)
  $\Phi$ = $\mathrm{true}$
  for $i \in [1\cdots n]$:
    $\exists \delta^1, \cdots, \delta^m : \phi_i$ = CorrectOne($T$,$\ptrace[t]$,$P$,$\correlt^i$)
    $\Phi$ = $\Phi \wedge (w^i = \phyper \veebar (w^i = 0 \wedge \phi_i))$ #\label{line:correctall:xor}#
  return $\exists \delta^1,\cdots,\delta^m,w^1,\cdots,w^n : \Phi$

def SRTR($T$, $P$, $\ptrace$,$\{\correlt^1, \cdots, \correlt^n\}$): #\label{line:srtr-start}#
  assert(CorrectAll($T$,$P$,$\ptrace$,$\{\correlt^1, \cdots, \correlt^n\}$))
  minimize($\Sigma_{i=1} w^i + \Sigma^m_{j=1} \|\delta^j\|$) #\label{line:srtr:minimize}#
  $\{ \pparam{x^1},\cdots,\pparam{x^m}\}$ = Rep($T$)
  return $[\pparam{x^1} \pmapsto
P(\pparam{x^1}) + \delta^1,\cdots, \pparam{x^m} \pmapsto  P(\pparam{x^m}) +
\delta^m]$  #\label{line:srtr-end}#

\end{lstlisting}
\caption{The core \technique{} algorithm.}
\figlabel{srtr-algo}
\end{figure}

\subsection{Residual Transition Functions}
\seclabel{program-analysis}
\lstset{language=algo}

The goal of \technique{} is to adjust the parameters such that for
each correction ($c_t$) the transition function produces the corrected
next-state instead of the actual state recorded at time $t+1$. \technique{} first
simplifies the problem by specializing the transition function using
the state, variables, and inputs recorded at time $t$.
We
call this simplified function the \emph{residual transition function}.
\figref{example-outputs} shows the residual transition function for the correction at
$t=5$.
Since the
state at this time-step (\treltstate{\tau_5}) is \state{Go To},
the residual transition function only has the code from the
branch that handles this case (\ie{} the code from lines
\ref{line:goto-start}--\ref{line:goto-end}). Furthermore,
we substitute the input and variable identifiers with
concrete values from the trace element and simplify expressions as
much as possible.
Therefore, the only identifiers that remain are parameters.
This approach is known as \emph{partial evaluation}~\cite{jones:pe}.
At a later step, \technique{} translates this residual transition function into a formula
for an SMT solver.

A potential problem with this approach is that SMT solvers do not have
decision procedures that support trigonometric functions, which
occur frequently in RSMs. Our example also
uses trigonometric functions in several expressions. Fortunately, most of these
trigonometric functions are applied to inputs and variables, thus they
are substituted with concrete values in the residual. For example,
line~\ref{line:trig-input} calculates
\lstinline|$\sin(\cpin{robotAng})$| and \lstinline|$\cos(\cpin{robotAng})$|,
but \lstinline|$\cpin{robotAng}$| is an input. Thus, the residual substitutes
the identifier with its value from the trace element
($\trelt_5.\cpin{robotAng}= 0$) and simplifies the trigonometric
expressions. In contrast, line~\ref{line:trig-param} applies a trigonometric
function to a parameter (\lstinline|$\sin(\cpparam{viewAng})$|).
This makes \lstinline|$\cpparam{viewAng}$| an
\emph{unrepairable parameter} that cannot appear in the
residual transition function. \technique{} substitutes unrepairable parameters
with their concrete values from the parameter map.

In general, the \lstinline|MakeResidual| function of \technique{}
(lines~\ref{line:residual}--\ref{line:residual-end} in \figref{srtr-algo})
takes a transition function ($T$), a trace element ($\tau_t$), and a parameter
map ($P$) and produces a residual transition function by partially
evaluating the transition function with respect to the trace element and
the unrepairable parameters. We use a simple dataflow analysis to
calculate the unrepairable parameters (\lstinline|Unrep($T$)|)
and a canonical partial evaluator (\lstinline|Peval|)~\cite{jones:pe}.

\subsection{Transition Repair as a \maxsmt{} Problem}
\seclabel{max-smt}

Given the procedure for calculating residual transition functions, \technique{}
proceeds in three steps. 1) It translates each correction $c_i$
into an independent formula $\phi_i$. A solution to $\phi_i$ corresponds
to parameter adjustments that satisfy the correction $c_i$.
Note however that no solution exists if $c_i$ cannot be satisfied. 2) It
combines the formulas $\phi_{1\cdots n}$ for all corrections $c_{1\cdots n}$
from the previous step into a single
formula $\Phi$ with independent penalties $w^i$ for each sub-formula $\phi_i$.
A solution to $\Phi$ corresponds to parameter adjustments that satisfy a subset of the
corrections. Any unsatisfiable corrections incur a penalty. 3)
Finally, it formulates a \maxsmt{} problem that minimizes the magnitude of
adjustments ($\|\delta^j\|$) to the parameters, and the penalty ($w^i$) of
violated sub-formulas.

The \lstinline|CorrectOne| function transforms a single correction into a
formula.
This function
1) calculates the residual transition
function (\figref{srtr-algo}, line~\ref{line:correctone:residual}), 2) gets the
repairable
parameters (line~\ref{line:correctone:rep}), and 3) produces
a formula (line~\ref{line:correctone-end}) with a variable for each
repairable parameter.
In our running example, the transition function has four parameters,
but, as explained in the previous section, the residual has only
three parameters since \cpparam{viewAng} is unrepairable.
Therefore, the formula that corresponds to this residual (\figref{example-outputs})
has three variables ($\delta^1$, $\delta^2$, and $\delta^3$).
Moreover, since the correction ($c_5$) requires the next-state to be
\state{Kick}, which only occurs when the residual takes the true-branch
(line 3 of the residual), the body of the formula is equivalent to
the conditional expression (lines 1--2), but with each parameter replaced
by the sum of its concrete value (from $P$) and its adjustment (a $\delta$).
For example, the formula replaces $\cpparam{aimMargin}$
by $\pi/50 + \delta^1$. Therefore, when $\delta^1 = 0$, the parameter
is unchanged.

The \lstinline|CorrectAll| function supports multiple corrections
and uses \lstinline|CorrectOne| as a subroutine.
 The function iteratively builds
a  conjunctive formula $\Phi$, where each clause has a distinct
penalty $w^i$ and two mutually exclusive cases: either
$w^i = 0$ thus the clause has no penalty and the adjustments to the parameters
satisfy the $i$th correction (line~\ref{line:correctall:xor}); or
a penalty is incurred ($w^i = \phyper$) and the $i$th correction is violated.
Thus $\phyper \in \reals{+}$ is a hyperparameter that induces a tradeoff between
satisfying
more corrections \vs{} minimizing the magnitude of the adjustments:
large values of $\phyper$ satisfy more corrections with larger adjustments,
whereas small values of $\phyper$ satisfy fewer corrections with smaller adjustments.
The final formula has $m$ real-valued variables $\delta^i$ for adjustments to
the corresponding $m$ repairable parameters $\pparam{x}^i$, and
$n$ discrete variables $w^j$ that represent the penalty of violating formula
$\phi_j$ corresponding to correction $\correlt^j$.

Our example (\figref{example-outputs}) has one correction and three repairable
parameters. Therefore,
\lstinline|CorrectAll|
produces a formula with four variables: a single penalty ($w^1$)
and the three adjustments discussed above ($\delta^{1}$, $\delta^{2}$,
and $\delta^{3}$). The formula is a single exclusive-or: either
the penalty is zero and formula is equivalent to the result of
\lstinline|CorrectOne| or the penalty is one and the result of
\lstinline|CorrectOne| is ignored.

Finally, the \lstinline|SRTR| function uses \lstinline|CorrectAll| as a
subroutine and invokes the \maxsmt{} solver. This function
1) adds the assertion returned by \lstinline|CorrectAll| to the solver,
2) directs the solver to minimize the sum of penalties and the sum of the
magnitude of parameter changes (line~\ref{line:srtr:minimize} in \figref{srtr-algo}),
and 3) returns a map from repairable parameter names to their new values.
In our example, the solver calculates that the minimum-cost
solution has $\delta^2 = 0.5$ with other variables set to zero.
\ie{} we can satisfy the correction by adjusting \cpparam{maxDist}
from $80$ to $80.5$.

In summary, \technique{} adjusts parameters to satisfy user-provided
corrections. It is not always possible to find an adjustment that satisfies all
corrections. Moreover, there is a tradeoff between making larger adjustments
and satisfying more corrections. Therefore, \technique{} uses a \maxsmt{} solver
to formulate the parameter adjustment problem. A limitation of this approach
is that a parameter may appear in a context cannot be expressed as a formula for
the \maxsmt{} solver. We use a simple analysis to determine which parameters
cannot be repaired.

\section{Evaluation}

We evaluate \technique{} in four ways.
\begin{inparaenum}[1)]
  \item  We compare \technique{} to an exhaustive search;
  \item We show how the number of corrections affect RSM performance
  and that \technique{} requires only a small number of corrections
  to perform well;
  \item Using three RSMs, we show that \technique{} does not over-fit and
  performs well in new scenarios; and
  \item We use \technique{} to improve the performance of a real-world
  robot.
\end{inparaenum}

We present three RSMs in this section and measure their success rates as follows.
The \emph{attacker} (\figref{example-model}) fills the main offensive role in
robot soccer. Its success
rate is the fraction of the test scenarios where it successfully
kicks the ball into the goal. The \emph{deflector} (\figref{deflection-fsm})
plays a supporting role in robot soccer, performing one-touch
passing~\cite{bruce2008cmdragons}. Its success rate is
the fraction of the test scenarios where it successfully deflects
the ball. The \emph{docker} (\figref{docking_fsm}) is a non-soccer behavior which drives a
differential drive robot to line up and dock with a charging station. Its success rate is the fraction of the test scenarios where
it successfully docks with the charging station.

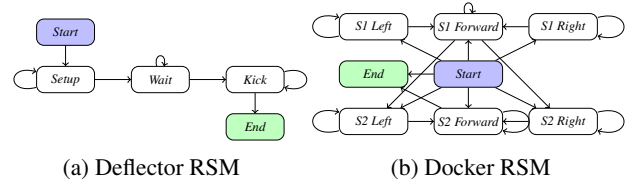
\begin{figure}
 \centering
 \begin{subfigure}[b]{0.45\linewidth}
 \centering
 \scalebox{.7}{%
    \begin{tikzpicture}[scale=0.151]
    \tikzstyle{every node}+=[rounded corners,minimum width=1.1cm,minimum height=0.5cm,node distance=0.9cm and 2cm,draw=black,inner sep=0pt,font=\scriptsize\itshape]
    \node[fill=blue!25] (start) {Start};
    \coordinate [right of=start] (rightstart);
    \node (setup) [below of=start] {Setup};
    \coordinate [right of=setup] (rightsetup);
    \node (wait) [right of=rightsetup] {Wait};
    \coordinate [right of=wait] (rightwait);
    \node (kick) [right of=rightwait] {Kick};
    \coordinate [right of=kick] (rightkick);
    \coordinate [left of=kick] (leftkick);
    \node[fill=green!25] (end) [below of=kick] {End};
    \path[every loop/.style={looseness=5}]
      (start) [->] edge (setup)
      (setup) [->] edge (wait)
      (wait) [->] edge (kick)
      (kick) [->] edge (end)
      (setup) [->] edge [loop left] (setup)
      (wait) [->] edge [loop above] (wait)
      (kick) [->] edge [loop right] (kick);
    \end{tikzpicture}
  }
  \caption{Deflector RSM}
  \figlabel{deflection-fsm}
     \end{subfigure}
     \begin{subfigure}[b]{0.52\linewidth}
     \centering
     \scalebox{.7}{%
      \begin{tikzpicture}[scale=0.151]
        \tikzstyle{every node}+=[rounded corners, minimum width=1.3cm,minimum height=0.5cm,node distance=0.9cm and 2cm,draw=black, inner sep=0pt,font=\scriptsize\itshape]
        \node[fill=blue!25] (start) {Start};
        \coordinate [left of=start] (leftstart);
        \node (s1forward) [above of=start] {S1 Forward};
        \coordinate [left of=s1forward] (left1forward);
        \node (s1left) [left of=left1forward] {S1 Left};
        \coordinate [right of=s1forward] (right1forward);
        \node (s1right) [right of=right1forward] {S1 Right};
        \node (s2forward) [below of=start] {S2 Forward};
        \coordinate [right of=s2forward] (right2forward);
        \coordinate [left of=s2forward] (left2forward);
        \node (s2left) [left of=left2forward] {S2 Left};
        \node (s2right) [right of=right2forward] {S2 Right};
        \node[fill=green!25] (end) [left of=leftstart] {End};
        \path[every loop/.style={looseness=5}]
          (start) [->] edge (s1forward)
          (start) [->] edge (s1left)
          (start) [->] edge (s1right)
          (start) [->] edge (s2forward)
          (start) [->] edge (s2left)
          (start) [->] edge (s2right)
          (start) [->] edge (end)
          (s1left) [->] edge (s1forward)
          (s1right) [->] edge (s1forward)
          (s1forward) [->] edge (s2left)
          (s1forward) [->] edge (s2right)
          (s2left) [->] edge (s2forward)
          (s2right) [->] edge (s2forward)
          (s2forward) [->] edge (end)
          (s1left) [->] edge [loop left] (s1left)
          (s2left) [->] edge [loop left] (s2left)
          (s1forward) [->] edge [loop above] (s2forward)
          (s2forward) [->] edge [loop right] (s2forward)
          (s1right) [->] edge [loop right] (s1right)
          (s2right) [->] edge [loop right] (s2right);
        \end{tikzpicture}
      }
    \caption{Docker RSM}
    \figlabel{docking_fsm}
     \end{subfigure}
     \caption{RSMs used for experiments}
   \figlabel{example_RSMs}
\end{figure}

\subsection{Comparison To Exhaustive Search}
\seclabel{parameter_search}

Using the attacker, we compare \technique{} to an exhaustive search to show
that 1) \technique{} is dramatically faster and 2) the adjustments found by
\technique{} are as good as those found by exhaustive search.
To limit the cost of exhaustive search, the experiment only repairs the
six parameters that affect transitions into the \state{Kick} state;
we bound the search space by the physical
limits of the parameters; and we discretize the resulting hypercube in
parameter space. We evaluate each parameter set using 13
simulated positions and manually specify if the position should
transition to the \state{Kick} state.

We evaluate the initial parameter values, the \technique{}-adjusted
parameters, and the parameters found by exhaustive search on
20,000 randomly generated scenarios. \tabref{comparison-results} reports
the success rate and running time of each approach. \technique{}
and exhaustive search achieve a comparable success rate. However,
\technique{} completes in 10 ms whereas exhaustive search takes 1,300 CPU
hours (using 100 cores).

\begin{table}
  \centering
  \setlength\tabcolsep{1.5pt} 
  \begin{tabular}{|l| r| c |}
  \hline
  \textbf{Method} &\textbf{Success Rate ($\%$)}  & \textbf{CPU Time} \\
  \hline
  Initial Parameters & 44 & ---  \\
  \hline
  Exhaustive Search & 89 & \hfill 1,300 hr \\
  \hline
  \technique{} & 89 & \hfill 10 ms \\
  \hline
  \end{tabular}
  \vspace{-.5em}
  \caption{Success rate and CPU time compared to exhaustive search.}
  \tablabel{comparison-results}
  \vspace{-.8em}
\end{table}

\subsection{Scalability}
\seclabel{performance}

Using the attacker, we now show how the number of corrections affects \technique{} solver
time and success rate.
Starting with the same set of initial parameters used in
\secref{parameter_search}, we first create a training dataset
by simulating the attacker in 40 randomly generated scenarios.
For each scenario, we provide one correction, thus we have a training
set of 40 corrections.
Each trial applies \technique{} to a subset of these corrections
and evaluates the success rate using 150 random test scenarios.
We repeat this procedure for each number of corrections $N \in [1,40]$ with 50
randomly chosen subsets of the training set (\ie{} 300,000 total trials).
\figref{corrVsSuc} shows how the number of corrections affects the success rate.
It is possible for a single informative correction to dramatically increase the
success rate, or for a particularly under-informative correction to have little
effect. Therefore we also report the mean success rate and show the $99\%$
confidence interval in gray. Success rate increases with the
number of corrections, and with
23 corrections, the attacker reaches a peak success rate of $87\%$.
\figref{time} shows that the solver time increases linearly with the number of
corrections, and remains less than 0.04s with 40 corrections.

\begin{figure}
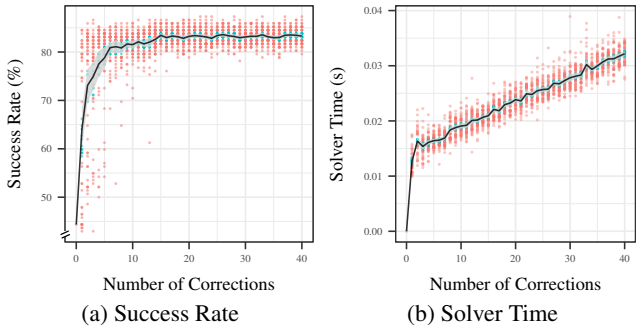

 \captionsetup[subfigure]{aboveskip=-1pt,belowskip=-4pt}
 \centering
 \begin{subfigure}[b]{0.49\linewidth}
    \centering
    \tikzexternalenable
    \tikzsetnextfilename{corrections_vs_success}
    \include{images/num_vs_succ}
    \tikzexternaldisable
    \vspace{-.6em}
    \caption{Success Rate}
    \figlabel{corrVsSuc}
 \end{subfigure}
 \begin{subfigure}[b]{0.49\linewidth}
    \centering
    \tikzexternalenable
    \tikzsetnextfilename{corrections_vs_time}
    \include{images/num_vs_time}
    \tikzexternaldisable
    \vspace{-.6em}
    \caption{Solver Time}
    \figlabel{time}
 \end{subfigure}
 \vspace{-.2em}
 \caption{Scaling of \technique{} with number of corrections. We report the
 mean as a line, the $99\%$ confidence interval in grey, inliers in blue,
 and outliers in red. Darker points
 represent more occurrences.}
 \vspace{-.3em}
\end{figure}

\subsection{Generalizability}
\seclabel{controller_repair}

Since \technique{} uses a handful of user-provided corrections to adjust
parameters, there is a risk that it may over-fit and underperform in new
scenarios. We use three RSMs to show
that 1) \technique{}-adjusted parameters generalize to new scenarios and that 2)
\technique{} outperforms a domain-expert who has 30 minutes to manually
adjust parameters.

\tabref{RSM_stats} summarizes the results of this experiment. We evaluate the
success rate of the Attacker, Deflector, and Docker RSMs on a test dataset with
several thousand test scenarios each. The baseline parameters that we use for
these RSMs have a low success rate. We give a domain expert complete access to
the RSM code (\ie{} the transition and emission functions), and subsequently
our simulator for 30 mins. In that time, the expert is able to dramatically
increase the success rate of the Deflector, but has minimal impact on the success
rate of the Attacker and the Docker. Finally, we apply \technique{} using a
handful of corrections and the baseline parameters. The
\technique{}-adjusted parameters perform significantly better than the baseline
and domain-expert parameters.

The heat maps in \figref{heatmap} illustrates how parameters found by the domain-expert,
and by \technique{} generalize to novel scenarios with the Attacker. In
both heat maps, the goal is the green bar and the initial position of the
Attacker is at the origin. Each coordinate corresponds to an initial position
of the ball and for each position we set the ball's initial velocity to 12
uniformly distributed
angles. With the expert-adjusted parameters, the Attacker performs well when
the ball starts in its immediate vicinity, but performs poorly otherwise.
However, with \technique{}-adjusted parameters, the Attacker is able to catch
or intercept the ball from most positions on the field.
For this result, we required only two corrections and the cross-marks in the
figure show the initial position of the ball for both corrections.
Therefore, although \technique{} only adjusted parameters to account for these two
corrections, the result generalized to many other positions on the field.

\begin{table}
  \setlength\tabcolsep{1.5pt} 
 \begin{tabular}{| l | r | r | r | r | r | r |}
  \hline
  \multirow{ 2}{*}{\textbf{RSM}} &
  \multirow{ 2}{*}{\textbf{Params}} &
  \multicolumn{1}{l|}{\textbf{\technique{}}}  &
  \multicolumn{1}{l|}{\multirow{ 2}{*}{\textbf{Tests}}}&
  \multicolumn{3}{c|}{\textbf{Success Rates ($\%$)}} \\
  \cline{5-7}
  & & \textbf{Corrections} & & \textbf{Baseline} & \textbf{Expert} &
\textbf{\technique} \\
  \hline
  Attacker    & 12 & 2 & 57,600 & 42 & 44 & 89 \\
  \hline
  Deflector  & 5 & 3 & 16,776 & 1 & 65 & 80 \\
  \hline
  Docker     & 9 & 3 & 5,000 & 0 & 0 & 100 \\
  \hline
  \end{tabular}
  \vspace{-.7em}
  \caption{Success rates for baseline, expert, and \technique{} parameters.}
  \tablabel{RSM_stats}
  \vspace{-.7em}
\end{table}

\begin{figure}
 \centering
 \begin{subfigure}[b]{0.49\linewidth}
    \centering
    \includegraphics[width=0.9\columnwidth]{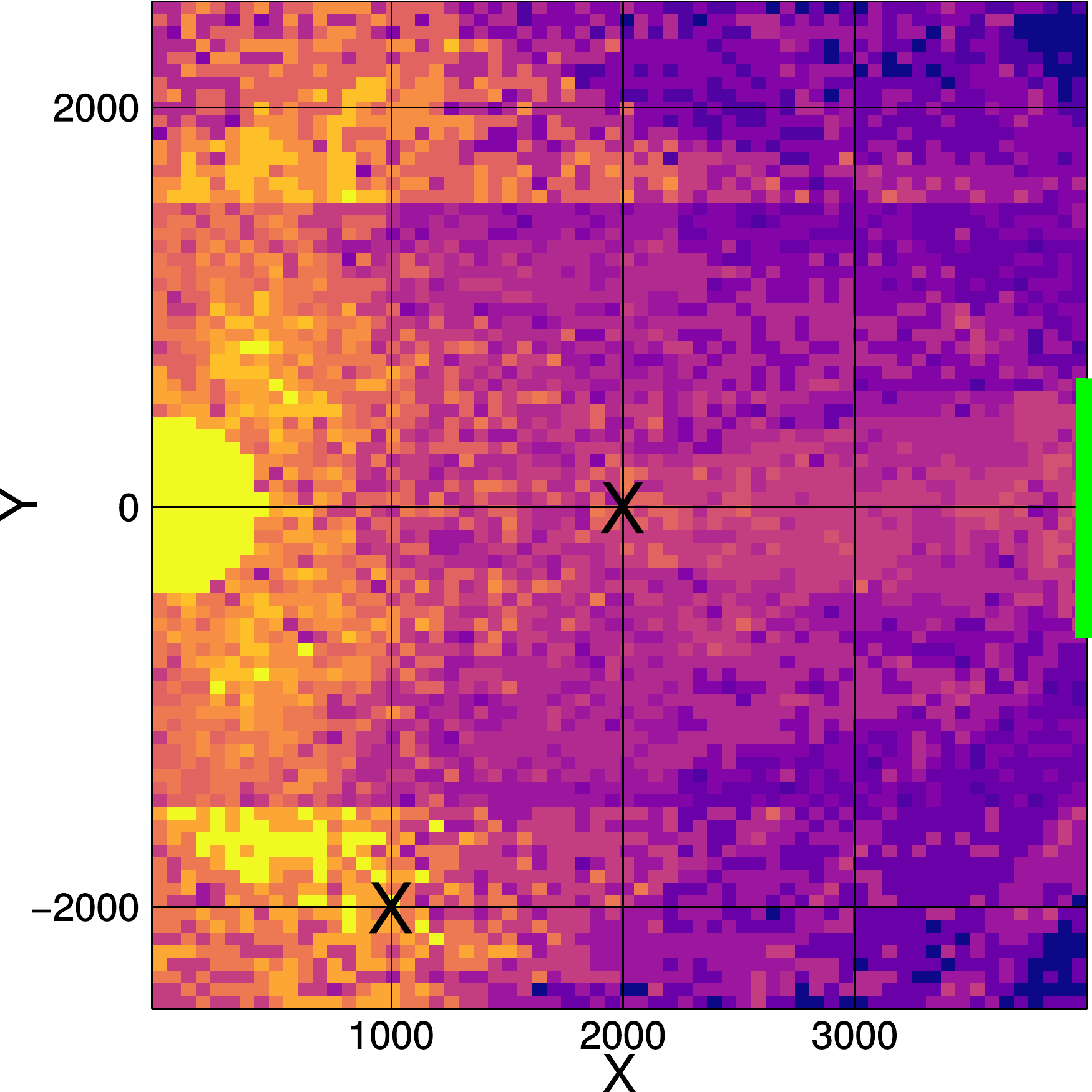}
    \caption{Expert-adjusted parameters.}
 \end{subfigure}
 \begin{subfigure}[b]{0.49\linewidth}
    \centering
    \includegraphics[width=0.9\columnwidth]{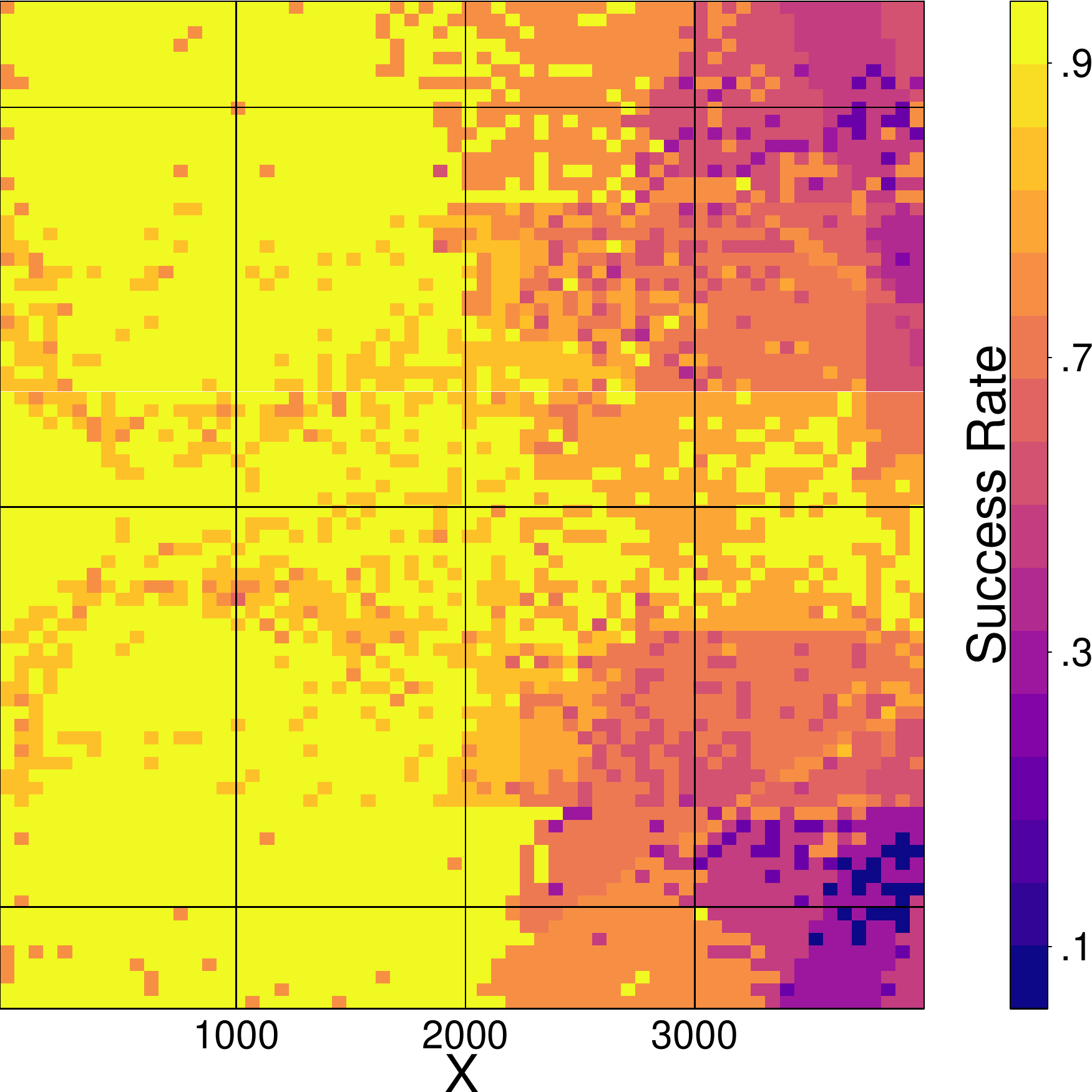}
    \caption{\technique{}-adjusted parameters.}
  \end{subfigure}
  \vspace{-.7em}
  \caption{Attacker success rate with respect to different initial ball
  positions. The corrections are marked with a cross.}
  \figlabel{heatmap}
\end{figure}

\begin{table}[tb]

  \centering
  \begin{tabular}{ | l | r |}
  \hline
  \textbf{Trial} & \textbf{Success Rate ($\%$)} \\
  \hline
  \cellcolor{blue!25} Competition Attacker & \cellcolor{blue!25} 75 \\
  \hline
  Parameters from Simulation & 24 \\
  \hline
  Real World \technique{} Tuning 1 & 73 \\
  \hline
  Real World \technique{} Tuning 2 & 85 \\
  \hline
  \end{tabular}
  \vspace{-.5em}
  \caption{Attacker success rates on a real robot.}
  \tablabel{real_tuning}
  \vspace{-1em}
\end{table}

\subsection{Case Study: \technique{} In The Real World}
\seclabel{real_world_repair}

To evaluate \technique{} in the real world, we follow the same procedure that
experts use (summarized in \tabref{real_tuning}): we develop the Attacker in a simulator, we adjust parameters
until it performs well in simulation, and then we find
that it performs poorly in the real world.
To evaluate the success rate of the attacker in the real world,
we start the ball from 18 positions on the field and repeat each position five times with the same velocity (\ie{} 90 trials).
The parameters from
simulation have a 25\% success rate. Using the
execution logs of this experiment, we apply
\technique{} with three corrections. The adjusted parameters increase the success rate
to 73\%. In practice, an expert would iteratively adjust parameters, so we
apply \technique{} again with 2 more corrections, which increases the success rate to 86\%.
Finally, our group has an attacker that we tested and optimized
extensively for RoboCup 2017, where it was part of a team that won the lower bracket.
This \emph{Competition Attacker} has additional states to handle special cases that do not occur in our tests. On our tests, the
Competition Attacker's success rate is 76\%. Therefore, with two iterations of
\technique{}, the simpler Attacker actually outperforms the Competition Attacker
in typical, real-world scenarios.

\section{Conclusion}
This paper presents SMT-based Robot Transition Repair (\technique{}),
which is a semi-automatic white-box approach for adjusting
the transition parameters of Robot State Machines.
We demonstrate that \technique{}:
\begin{inparaenum}[1)]
\item increases success rate for multiple behaviors;
\item finds new parameters quickly using a small number of annotations;
\item produces solutions which generalize well to novel situations; and
\item improves performance in a real world robot soccer application.
\end{inparaenum}

\section*{Acknowledgements}

This work is supported in part by AFRL and DARPA agreement \#FA8750-16-2-0042, and by
NSF grant CCF-1717636.

\bibliographystyle{named}
\bibliography{references}

\def\icra{ICRA}\def\iros{IROS}\def\rss{RSS}\def\aamas{AAMAS}\def\ijcai{IJCAI}\def\ase{ASE}\def\icse{ICSE}
\begin{thebibliography}{}

\bibitem[\protect\citeauthoryear{Abbeel and Ng}{2011}]{abbeel2011inverse}
Pieter Abbeel and Andrew~Y Ng.
\newblock Inverse reinforcement learning.
\newblock In {\em Encyclopedia of machine learning}, pages 554--558. Springer,
  2011.

\bibitem[\protect\citeauthoryear{Argall \bgroup \em et al.\egroup
  }{2009}]{argall2009survey}
Brenna~D. Argall, Sonia Chernova, Manuela Veloso, and Brett Browning.
\newblock {A Survey of Robot Learning From Demonstration}.
\newblock {\em Robotics and Autonomous Systems}, pages 469 -- 483, 2009.

\bibitem[\protect\citeauthoryear{Bai and Russell}{2017}]{bai2017efficient}
Aijun Bai and Stuart Russell.
\newblock {Efficient Reinforcement Learning with Hierarchies of Machines by
  Leveraging Internal Transitions}.
\newblock In {\em \icra}, 2017.

\bibitem[\protect\citeauthoryear{Biswas \bgroup \em et al.\egroup
  }{2014}]{biswas2014opponent}
Biswas, Mendoza, Zhu, Choi, Klee, and Veloso.
\newblock Opponent-driven planning and execution for pass, attack, and defense
  in a multi-robot soccer team.
\newblock In {\em \aamas}, pages 493--500, 2014.

\bibitem[\protect\citeauthoryear{Bj{\o}rner \bgroup \em et al.\egroup
  }{2015}]{bjorner2015nuz}
Nikolaj Bj{\o}rner, Anh-Dung Phan, and Lars Fleckenstein.
\newblock {$\nu$Z-an optimizing SMT solver}.
\newblock In {\em International Conference on Tools and Algorithms for the
  Construction and Analysis of Systems}, pages 194--199, 2015.

\bibitem[\protect\citeauthoryear{Bruce \bgroup \em et al.\egroup
  }{2008}]{bruce2008cmdragons}
James Bruce, Stefan Zickler, Mike Licitra, and Manuela Veloso.
\newblock {CMDragons: Dynamic Passing and Strategy on a Champion Robot Soccer
  Team}.
\newblock In {\em \icra}, pages 4074--4079, 2008.

\bibitem[\protect\citeauthoryear{Cano \bgroup \em et al.\egroup
  }{2016}]{cano2016automatic}
Jos\'e Cano, Alejandro Bordallo, Vijay Nagarajan, Subramanian Ramamoorthy, and
  Sethu Vijayakumar.
\newblock {Automatic Configuration of ROS Applications for Near-Optimal
  Performance}.
\newblock In {\em \iros}, pages 2217--2223, 2016.

\bibitem[\protect\citeauthoryear{Dantam \bgroup \em et al.\egroup
  }{2013}]{dantam2013}
Neil Dantam, Ayonga Hereid, Aaron Ames, and Mike Stilman.
\newblock {Correct Software Synthesis for Stable Speed-Controlled Robotic
  Walking}.
\newblock In {\em \rss}, 2013.

\bibitem[\protect\citeauthoryear{Devlin \bgroup \em et al.\egroup
  }{2017}]{devlin:robustfull}
Jacob Devlin, Jonathan Uesato, Surya Bhupatiraju, Rishabh Singh, Abdelrahman
  Mohammad, and Pushmeet Kohli.
\newblock {RobustFill}: Neural program learning under noisy {I/O}.
\newblock In {\em ICML}, 2017.

\bibitem[\protect\citeauthoryear{Gulwani}{2011}]{gulwani2011automating}
Sumit Gulwani.
\newblock Automating string processing in spreadsheets using input-output
  examples.
\newblock In {\em ACM SIGPLAN Notices}, pages 317--330. ACM, 2011.

\bibitem[\protect\citeauthoryear{Holtz and Biswas}{2017}]{holtz2017automatic}
Jarrett Holtz and Joydeep Biswas.
\newblock {Automatic Extrinsic Calibration of Depth Sensors with Ambiguous
  Environments and Restricted Motion}.
\newblock In {\em \iros}, pages 2235--2240, 2017.

\bibitem[\protect\citeauthoryear{Jones \bgroup \em et al.\egroup
  }{1993}]{jones:pe}
Neil~D. Jones, Carsten~K. Gomad, and Peter Sestoft.
\newblock {\em Partial Evaluation and Automatic Program Generation}.
\newblock Prentice-Hall, Inc., 1993.

\bibitem[\protect\citeauthoryear{Kalm{\'a}r-Nagy \bgroup \em et al.\egroup
  }{2004}]{kalmar2004near}
Tam{\'a}s Kalm{\'a}r-Nagy, Raffaello D’Andrea, and Pritam Ganguly.
\newblock {Near-Optimal Dynamic Trajectory Generation and Control of an
  Omnidirectional Vehicle}.
\newblock {\em Robotics and Autonomous Systems}, pages 47--64, 2004.

\bibitem[\protect\citeauthoryear{Kamar}{2016}]{Kamar2016DirectionsIH}
Ece Kamar.
\newblock {Directions in Hybrid Intelligence: Complementing AI Systems with
  Human Intelligence}.
\newblock In {\em \icra}, 2016.

\bibitem[\protect\citeauthoryear{Mechtaev \bgroup \em et al.\egroup
  }{2015}]{mechtaev:directfix}
Sergey Mechtaev, Jooyong Yi, and Abhik Roychoudhury.
\newblock {DirectFix}: {Looking for Simple Program Repairs}.
\newblock In {\em \icse}, pages 448--458, 2015.

\bibitem[\protect\citeauthoryear{Meri{\c{c}}li \bgroup \em et al.\egroup
  }{2012}]{mericli2012multi}
{\c{C}}etin Meri{\c{c}}li, Manuela Veloso, and H~Levent Ak{\i}n.
\newblock Multi-resolution corrective demonstration for efficient task
  execution and refinement.
\newblock {\em International Journal of Social Robotics}, pages 423--435, 2012.

\bibitem[\protect\citeauthoryear{Mericli \bgroup \em et al.\egroup
  }{2014}]{mericli2014interactive}
Cetin Mericli, Steven~D. Klee, Jack Paparian, and Manuela Veloso.
\newblock {An Interactive Approach for Situated Task Specification Through
  Verbal Instructions}.
\newblock In {\em \aamas}, pages 1069--1076, 2014.

\bibitem[\protect\citeauthoryear{Miyazawa \bgroup \em et al.\egroup
  }{2017}]{miyazawa2017}
Alvaro Miyazawa, Pedro Ribeiro, Wei Li, Ana Cavalcanti, and Jon Timmis.
\newblock Automatic property checking of robotic applications.
\newblock In {\em \iros}, pages 3869--3876, 2017.

\bibitem[\protect\citeauthoryear{Nashed and Biswas}{2018}]{nashed2018human}
Samer Nashed and Joydeep Biswas.
\newblock {Human-in-the-Loop SLAM}.
\newblock In {\em AAAI}, 2018.

\bibitem[\protect\citeauthoryear{Nedunuri \bgroup \em et al.\egroup
  }{2014}]{nedunuri2014}
Srinivas Nedunuri, Sailesh Prabhu, Mark Moll, Swarat Chaudhuri, and Lydia~E.
  Kavraki.
\newblock {SMT-based synthesis of integrated task and motion plans from plan
  outlines}.
\newblock In {\em \icra}, pages 655--662, 2014.

\bibitem[\protect\citeauthoryear{Reiter}{1987}]{reiter1987theory}
Raymond Reiter.
\newblock {A Theory of Diagnosis from First Principles}.
\newblock {\em Artificial Intelligence}, pages 57--95, 1987.

\bibitem[\protect\citeauthoryear{Weiss \bgroup \em et al.\egroup
  }{2017}]{tortoise:weiss}
Aaron Weiss, Arjun Guha, and Yuriy Brun.
\newblock {Tortoise: Interactive System Configuration Repair}.
\newblock In {\em \ase}, pages 625--636, 2017.

\bibitem[\protect\citeauthoryear{Wong \bgroup \em et al.\egroup
  }{2014}]{wong2014correct}
Kai~Weng Wong, R{\"a}diger Ehlers, and Hadas Kress-Gazit.
\newblock {Correct High-level Robot Behavior in Environments with Unexpected
  Events}.
\newblock In {\em \rss}, 2014.

\end{thebibliography}

\end{document}